\documentclass[11pt,english]{article}
\usepackage[T1]{fontenc}
\usepackage[latin1]{inputenc}
\usepackage{babel}
\usepackage{setspace}
\makeatletter

\newcommand{\LyX}{L\kern-.1667em\lower.25em\hbox{Y}\kern-.125emX\spacefactor1000}

\newcommand{\lyxaddress}[1]{
  \par {\raggedright #1 
  \vspace{1.4em}
  \noindent\par}
}

\makeatother

\begin{document}

\begin{spacing}{1.00}

\title{Boosting the Differences : A Fast Bayesian Classifier Neural Network }
\end{spacing}

\begin{spacing}{1.00}

\author{Ninan Sajeeth Philip\thanks{
E-mail: nsp@stthom.ernet.in
} \protect\( \, \, \protect \)and K. Babu Joseph \thanks{
E-mail:smanager@giasmd01.vsnl.net.in
}}
\end{spacing}

\maketitle
\begin{spacing}{1.00}

\lyxaddress{Department of Physics, Cochin University of Science and Technology, Kochi,
India.}
\end{spacing}

\begin{abstract}
\begin{spacing}{1.00}
A new classifier based on Bayes' principle that assumes the clustering of attribute
values while boosting the attribute differences is presented. The method considers
the error produced by each example in the training set in turn and upweights
the weight associated to the probability \( P(U_{m}\mid C_{k}) \) of each attribute
of that example. In this process the probability density of identical attribute
values flattens out and the differences get boosted up. Using four popular datasets
from the UCI repository, some of the characteristic features of the network
are illustrated. The network is found to have optimal generalization ability
on all the datasets. For a given topology, the network converges to the same
classification accuracy and the training time as compared to other networks
is less. It is also shown that the network architecture is suitable for parallel
computation and that its optimization may also be done in parallel. 


\newpage\end{spacing}

\end{abstract}

\section{Introduction}

Machine learning has been one of the most active areas of research in recent
times. A significant boost to this was the introduction of the backpropagation
algorithm by Rumelhart \cite{Rumelhart86, Rumelhart93}. Another class of learning
algorithms based on the Bayesian theory also became popular in the search for
better learning algorithms. Standard Bayesian networks involves a lot of computational
labour so that they are used only in highly demanding situations. In many practical
problems, a simplified version of the Bayesian classifiers known as the 'naive'
Bayesian (NB) classifier \cite{Papert69} performs equally well. The basis of
the Bayesian classifier is that identical examples will cluster together in
an \( n \) dimensional attribute space (parameter space or feature space) and
that it may be efficiently classified presuming that the attributes are independent
for a given class of the example. If we consider a discrete space \( \Re ^{n} \)
to represent an example, the resolution with which the attributes should be
mapped to identify the clusters is determined on the basis of the separability
of the classes. If the attributes have narrow margins, a high resolution is
required. For example, a particular color might be unique to an object. Then
the resolution is not very significant in classifying the object. Obviously,
the likelihood for overlapping attribute values will be distributed among the
classes. Assuming that the resolution is reasonably chosen, for each class we
generate a table of the likelihoods for each discrete attribute value. If the
attribute values are continuous, we construct bins of uniform width to hold
them. This is equivalent to digitizing the data with some suitable step size.
The independence of attribute values require that the product of the corresponding
values of the likelihoods of the attributes be an extremum for the respective
class of the example. 

However, one major drawback of the NB classifier is that it is not able to learn
the classical XOR problem. In fact, it fails on any problem that does not support
the independence of the attribute values. In this connection, Elkan \cite{Elkan97}
showed that the introduction of noise can improve the ability of naive Bayesian
classifiers when dealing with XOR kind of problems. He further argued that addition
of noise has the effect that the 'true concept resembles a disjunction of conjunctions
with noise'. Here we introduce a slightly different approach. We apply certain
empirical rules on the attributes during the training process and use this information
to do the classification later. A simple boosting algorithm is then used to
amplify the difference produced by these rules to improve the classification
accuracy. Some examples from the UCI repository data sets for machine learning
are used to benchmark the proposed algorithm with other known methods for classification.

\section{Naive Bayesian learning}

A convincing formalism on how the degrees of belief should vary on the basis
of evidence is obtained from Bayesian theory. If \( P(U_{i}\mid H) \) represent
the likelihood by which the evidence (feature value ) \( U_{i} \) occurs in
the hypothesis (class) \( H \), which itself has a probability of occurrence
\( P(H) \), then Bayes' rule states the posterior probability, or the degree
of belief with which this evidence \( U_{i} \) would propose the occurrence
of the hypothesis \( H \) as:

\[
P(H\mid U_{i})=\frac{P(U_{i}\mid H)P(H)}{\sum _{i}P(U_{i}\mid H)P(H)}\]
Assume that we have a dataset with N training examples represented as \( S=\{x_{1},x_{2},......x_{N}\} \).
Each of these examples \( x_{n} \) is represented by a set of \( M \) independent
attribute values. Thus we represent \( x \) as : 
\[
x\equiv U_{1}\odot U_{2}\odot .......U_{M}\]
 where \( \odot  \) denotes the logical AND operation. This has the analogy
to the statement that Rose is an object with red color AND soft petals AND striking
smell. We further assume that the training set is complete with \( K \) different
known discrete classes. A statistical analysis should assign a maximal value
of the conditional probability \( P(C_{k}\mid U) \) for the actual class \( C_{k} \)
of the example. By Bayes' rule this probability may be computed as :
\[
P(C_{k}\mid U)=\frac{P(U\mid C_{k})\, P(C_{k})}{\sum _{K}P(U\mid C_{k})\! P(C_{k})}\]

\( P(C_{k}) \) is also known as the background probability. Since the attributes
are associated to the example vector by a logical AND condition, \( P(U\mid C_{k}) \)
is given by the product of the probabilities due to individual attributes. Thus,

\[
P(U\mid C_{k})=\prod _{m}P(U_{m}\mid C_{k})\]
 Following the axioms of set theory, one can compute \( P(U_{m}\mid C_{k}) \)
as \( P(U_{m}\cap C_{k}) \). This is nothing but the ratio of the total count
of the attribute value \( U_{m} \) in class \( C_{k} \) to the number of examples
in the entire training set. Thus naive Bayesian classifiers complete a training
cycle much faster than perceptrons or feed-forward neural networks. Elkan reports
a time of less than a minute to learn a data set of 40,000 examples each with
25 attribute values on a 'modern workstation'.

\section{Boosting\label{boosting} }

Boosting is an iterative process by which the network upweights misclassified
examples in a training set until it is correctly classified. The Adaptive Boosting
(AdaBoost) algorithm of Freund and Schapire \cite{Freund95, Freund97} attempts
the same thing. In this paper, we present a rather simple algorithm for boosting.
The structure of our network is identical to AdaBoost in that it also modifies
a weight function. Instead of computing the error in the classification as the
total error produced in the training set, we take each misclassified example
and apply a correction to its weight based on its own error. Also, instead of
upweighting an example, our network upweights the weight associated to the probability
\( P(U_{m}\mid C_{k}) \) of each attribute of the example. Thus the modified
weight will affect all the examples that have the same attribute value even
if its other attributes are different. During the training cycle, there is a
competitive update of attribute weights to reduce the error produced by each
example. It is expected that at the end of the training epoch the weights associated
to the probability function of each attribute will stabilize to some value that
produces the minimum error in the entire training set. Identical feature values
compete with each other and the differences get boosted up. Thus the classification
becomes more and more dependent on the differences rather than on similarities.
This is analogous to the way in which the human brain differentiates between
almost similar objects by sight, like for example, rotten tomatoes from a pile
of good ones. 

Let us consider a misclassified example in which \( P_{k} \) represent the
computed probability for the actual class \( k \) and \( P_{k}^{*} \) that
for the wrongly represented class. Our aim is to push the computed probability
\( P_{k} \) to some value greater than \( P^{*}_{k} \). In our network, this
is done by modifying the weight associated to each \( P(U_{m}\mid C_{k}) \)
of the misclassified item by the negative gradient of the error, i.e. \( \Delta W_{m}=\alpha \left[ 1-\frac{P_{k}}{P^{*}_{k}}\right]  \).
Here \( \alpha  \) is a constant which determines the rate at which the weight
changes. The process is repeated until all items are classified correctly or
a predefined number of rounds completes.

\section{The classifier network.}

Assuming that the occurrences of the classes are equally probable, we start
with a flat prior distribution of the classes ,i.e. \( P(C_{k})=\frac{1}{N} \).
This might appear unrealistic, since this is almost certain to be unequal in
most practical cases. The justification is that since \( P(C_{K}) \) is also
a weighting function, we expect this difference also to be taken care of by
the connection weights during the boosting process. The advantage on the otherhand
is that it avoids any assumptions on the training set regarding the prior estimation.
Now, the network presented in this paper may be divided into three units. The
first unit computes the Bayes' probability for each of the training examples.
If there are \( M \) number of attributes with values ranging from \( m_{min} \)
to \( m_{max} \) and belonging to one of the \( K \) discrete classes, we
first construct a grid of equal sized bins for each \( k \) with columns representing
the attributes and rows their values. Thus a training example \( S_{i} \) belonging
to a class \( k \) and having one of its attributes \( l \) with a value \( m \)
will fall into the bin \( B_{klm} \) for which the Euclidean distance between
the center of the bin and the attribute value is a minimum. The number of bins
in each row should cover the range of the attributes from \( m_{min} \) to
\( m_{max} \). It is observed that there exist an optimum number of bins that
produce the maximum classification efficiency for a given problem. For the time
being, it is computed by trial and error. Once this is set, the training process
is simply to distribute the examples in the training sets into their respective
bins. After this, the number of attributes in each bin \( i \) for each class
\( k \) is counted and this gives the probability \( P(U_{m}\mid C_{k}) \)
of the attribute \( m \) with value \( U_{m}\equiv i \) for the given \( C_{k}=k \).
The basic difference of this new formalism with that of the popular gradient
descent backpropagation algorithm and similar Neural Networks is that, here
the distance function is the distance between the probabilities, rather than
the feature magnitudes. Thus the new formalism can isolate overlapping regions
of the feature space more efficiently than standard algorithms. 

As mentioned earlier, this learning fails when the data set represent an XOR
like feature. To overcome this, associated to each row of bins of the attribute
values we put a tag that holds the minimum and maximum values of the other attributes
in the data example. This tag acts as a level threshold window function. In
our example, if an attribute value in the example happens to be outside the
range specified in the tag, then the computed \( P(U_{m}\mid C_{k}) \) of that
attribute is reduced to one-forth of its actual value (gain of 0.25). Applying
such a simple window enabled the network to handle the XOR kind of problems
such as the monk's problem efficiently.

The second unit in the network is the gradient descent boosting algorithm. To
do this, each of the probability components \( P(U_{m}\mid C_{k}) \) is amplified
by a connection weight before computing \( P(U\mid C_{k}) \). Initially all
the weights are set to unity. For a correctly classified example, \( P(U\mid C_{k}) \)
will be a maximum for the class specified in the training set. For the misclassified
items, we increment its weight by a fraction \( \Delta W_{m} \). For the experiments
cited here, the value of \( \alpha  \) was taken as 2.0. The training set is
read repeatedly for a few rounds and in each round the connection weights of
the misclassified items are incremented by \( \Delta W_{m}=\alpha \left[ 1-\frac{P_{k}}{P^{*}_{k}}\right]  \)
as explained in section \ref{boosting}, until the item is classified correctly. 

The third unit computes \( P(C_{k}\mid U) \) as :

\[
P(C_{k}\mid U)=\frac{\prod _{m}P(U_{m}\mid C_{k})\! W_{m}}{\sum _{K}\prod _{m}P(U_{m}\mid C_{k})\! W_{m}}\]
 If this is a maximum for the class given in the training set, the network is
said to have learned correctly. The wrongly classified items are re-submitted
to the boosting algorithm in the second unit.

\section{Experimental results}

For the experimental verification of the network model, we selected four well
known databases from the UCI repository. The first three examples are realistic
data on breast cancer, hypothyroid and diabetes patients, while the fourth one
is a set of three artificial datasets generated by Sebastian Thrun in 1991 for
benchmarking many popular learning algorithms. Each dataset illustrates some
characteristic feature of the naive Bayesian classifier proposed in this paper.

\subsection{Wisconsin breast cancer databases}

The Wisconsin breast cancer database represents a reasonably complex problem
with 9 continuous input attributes and two possible output classes. This data
set was donated by Dr. William H. Wolberg of the University of Wisconsin Hospitals.
The dataset consists of 683 instances and we divided it into a training set
of 341 examples and a test set of 342 examples each. The problem is to find
if the evidences indicates a Benign or Malignant neoplasm. Wolberg \cite{Wolberg90}
used 369 instances of the data (available at that point in time) for classification
and found that two pairs of parallel hyperplanes are consistent with 50\% of
the data. Accuracy on remaining 50\% of dataset was 93.5\%. It is also reported
that three pairs of parallel hyperplanes were found to be consistent with 67\%
of data and the accuracy on remaining 33\% was 95.9\%

The input attributes are: 

\vspace{0.3cm}
{\centering \begin{tabular}{|l|c|}
\hline 
Attribute&
Type\\
\hline 
\hline 
Clump Thickness&
continuous\\
\hline 
Uniformity of Cell Size&
continuous\\
\hline 
Uniformity of Cell Shape&
continuous\\
\hline 
Marginal Adhesion&
continuous\\
\hline 
Single Epithelial Cell Size&
continuous\\
\hline 
Bare Nuclei&
continuous\\
\hline 
Bland Chromatin&
continuous\\
\hline 
Normal Nucleoli&
continuous\\
\hline 
Mitoses&
continuous\\
\hline 
\end{tabular}\par}
\vspace{0.3cm}

Taha and Ghosh used all the 683 instances to test a hybrid symbolic-connectionist
system \cite{Ghosh96}. Using a Full Rule Extraction algorithm, they report a
recognition rate of 97.77\%. The proposed network using 7 bins for each attribute
converged in 87 iterations to produce a classification accuracy of 97.95 \%
on the independent test set. Only seven out of 342 examples were misclassified.

\subsection{Thyroid databases}

The thyroid database was donated by Randolf Werner in 1992. It consists of 3772
learning examples and 3428 testing examples readily usable as ANN training and
test sets. In the repository, these datasets are named : pub/machine-learning-databases/thyroid-disease/ann{*}.
Each example has 21 attributes, 15 of which are binary and 6 are continuous.
The problem is to determine whether a patient referred to the clinic is hypothyroid.
The output from the network is expected to be one of the three possible classes,
namely: (i) normal (not hypothyroid), (ii) hyperfunction and (iii) subnormal
function. In the dataset, 92 percent of the patients are not hyperthyroid and
thus any reasonably good classifier should have above 92\% correct predictions.
Schiffmann \textit{et al.,} \cite{Schiffmann94} used this dataset to benchmark
15 different algorithms. Fourteen of the networks had a fixed topology of 3
layers with 21 input nodes, 10 hidden nodes and 3 output nodes. The network
was fully interconnected. The other network used was a cascade correlation network
with 10 and 20 units each. Using a SPARC2 CPU, the reported training time on
the dataset varied from 12 to 24 hours. On the otherhand, using an ordinary
Celeron 300MHz Linux PC, our network using 9 bins each for the continuous attributes
and 2 bins each for the binary attributes took less than 10 minutes to attain
a classification accuracy better than the best results reported by Schiffmann
\textit{et al} . We summarize their best results along with ours in table I.

\vspace{0.3cm}
{\centering \begin{tabular}{|l|c|c|}
\hline 
Algorithm&
Training Set &
Test set \\
\hline 
\hline 
Backprop&
99.13&
97.58\\
\hline 
Backprop(batch mode)&
92.63&
92.85\\
\hline 
Backprop(batch mode)+ Eaton and Oliver&
92.47&
92.71\\
\hline 
Backprop+Darken and Moody&
99.20&
97.90\\
\hline 
J. Schmidhuber&
98.36&
97.23\\
\hline 
R.Salomon&
94.64&
94.14\\
\hline 
Chan and Fallside&
94.67&
94.17\\
\hline 
Polak-Ribiere+line search&
94.70&
94.17\\
\hline 
Conj. gradient + line search&
94.57&
93.84\\
\hline 
Silva and Almeida&
\textbf{\underbar{99.60}}&
\textbf{\underbar{98.45}}\\
\hline 
SuperSAB&
99.55&
98.42\\
\hline 
Delta-Bar-Delta&
99.20&
98.02\\
\hline 
RPROP&
99.58&
98.02\\
\hline 
Quickprop&
99.60&
98.25\\
\hline 
Cascade correlation 10 units&
99.84&
98.42\\
\hline 
Cascade correlation 20 units&
\textbf{\underbar{100}}&
\textbf{\underbar{98.48}}\\
\hline 
Our network&
\textbf{\underbar{99.0}}6&
\textbf{\underbar{98.60}}\\
\hline 
\end{tabular}\par}
\vspace{0.3cm}

\noindent Table I : A comparison of the efficiency of the proposed classifier
with other networks on the UCI Thyroid database. 
\medskip{}

\subsection{Pima Indians Diabetes Database}

The Pima Indian diabetes database, donated by Vincent Sigillito, is a collection
of medical diagnostic reports of 768 examples from a population living near
Phoenix, Arizona, USA. The paper dealing with this data base \cite{Smith88}
uses an adaptive learning routine that generates and executes digital analogs
of perceptron-like devices, called ADAP. They used 576 training instances and
obtained a classification of 76 \% on the remaining 192 instances. The samples
consist of examples with 8 attribute values and one of the two possible outcomes,
namely whether the patient is ``tested positive for diabetes\char`\"{} (indicated
by output one) or not (indicated by two). The database now available in the
repository is a refined version by George John in October 1994 and it has 512
examples in the training set and 256 examples in the test set. The attribute
vectors of these examples are:.

\vspace{0.3cm}
{\centering \begin{tabular}{|l||c|}
\hline 
Attribute&
Type\\
\hline 
\hline 
Number of times pregnant&
continuous\\
\hline 
Plasma glucose concentration&
continuous\\
\hline 
Diastolic blood pressure (mm Hg)&
continuous\\
\hline 
Triceps skin fold thickness (mm)&
continuous\\
\hline 
2-Hour serum insulin (mu U/ml)&
continuous\\
\hline 
Body mass index {[}weight in kg/(height in m)\( ^{2} \){]}&
continuous\\
\hline 
Diabetes pedigree function&
continuous\\
\hline 
Age (years)&
continuous\\
\hline 
\end{tabular}\par}
\vspace{0.3cm}

We use this dataset to illustrate the effect of the topology (in terms of the
number of bins per attribute) on the generalization ability of the proposed
network. With a twelve fold cross-validation and special pre-processing, the
test result reported for the data by George John is 77.7 using the LogDisc algorithm.
Table II summarizes the generalization obtained on our network for the same
dataset without any pre-processing. The first column indicates the number of
bins used for each attribute and is followed by the classification success percentage
for the training and test sets.

\vspace{0.3cm}
{\centering \begin{tabular}{|c|l|c|c|}
\hline 
Sl. No.&
No. of bins for each attribute&
Training data&
Test data\\
\hline 
\hline 
1&
5-5-5-5-14-5- 5-5&
82.42 \%&
72.66 \%\\
\hline 
2&
5-5-5-5-5-30-5-5&
82.42 \%&
72.66 \%\\
\hline 
3&
5-5-5-5-14-30-5-5&
84.57 \%&
75.00 \%\\
\hline 
4&
8-5-5-5-14-30-5-5&
83.79 \%&
76.17 \%\\
\hline 
5&
8-5-5-5-14-30-5-6&
84.77 \% &
76.95 \%\\
\hline 
\end{tabular}\par}
\vspace{0.3cm}

\noindent Table II : The table illustrates how the optimization of the bins
may be done for the proposed network on parallel architecture.
\medskip{}

As it can be seen, the optimal topology is (8-5-5-5-14-30-5-6) giving a classification
accuracy of 76.95\%. It may also be noted that the process of optimization of
bin number obeys additive property. Thus when attributes five and six uses 14
and 30 bins each, the resulting accuracy is 75 \% which is about 2.5 \% above
that produced by them individually. This means that the optimization of the
topology of the network may be automated in parallel on a virtual machine to
make the best possible network for a task. Since naive Bayesian networks also
support parallel computation of attribute values, this network is well suited
for parallel architecture producing high throughput. Future research shall explore
this possibility in detail.

\subsection{Monks' problems}

In 1991 Sebastian Thrun composed these datasets for benchmarking many known
learning algorithms. These datasets were generated using propositional formulas
over six attributes. The formulas used for each dataset are: (1) \( A_{1}=A_{2}=a_{51} \)
(2) exactly two of \( A_{1} \)through \( A_{6} \) have their first value,
and (3) {[}\( A_{5}=a_{53} \) and \( A_{4}=a_{41} \){]} or {[}\( A_{5}\neq a_{54} \)
and \( A_{2}\neq a_{23} \){]}. In addition to this, the third dataset was added
with 5\% classification noise also. All the three data sets have an XOR flavor
by design. Using a naive Bayesian classifier with AdaBoost algorithm, Elkan
reports that the classifier succeeded only on the third dataset. This is qualitatively
explained \cite{Elkan97} as caused by the inherent inability of boosting algorithms
to boost a dataset unless the intermediate hypothesis, including the first,
has a probability greater than 50 \%. In XOR kind of problems, this is exactly
50\% and no boosting is possible. With the addition of noise this situation
changes and the boosting algorithm takes advantage of it to produce better classification.

In the proposed network, we used a threshold window function tagged to each
possible attribute value of the training data. The purpose of this function
is to bring in some difference in the computed probability values of the naive
Bayesian classifier that may be used by the boosting algorithm to differentiate
the classes. A uniform size of 4 bins were used to represent each attribute.
We summarize the results in table III. The number of examples in the data are
shown in brackets and the other numbers shown are the number of correctl`y classified
examples in class 1 and class 2 respectively.

\vspace{0.3cm}
{\centering \begin{tabular}{|l|l|}
\hline 
Dataset&
Our Network\\
\hline 
\hline 
Monk's training set 1 (124)&
62, 62 : 100 \%\\
\hline 
Monk's test set 1 (432)&
214, 205 : 96.99 \%\\
\hline 
Monk's training set 2 (169)&
17, 95 : 66.27 \%\\
\hline 
Monk's test set 2 (432)&
38, 252 : 67.23 \%\\
\hline 
Monk's training set 3 (122)&
57, 61 : 96.72 \%\\
\hline 
Monk's test set 3 (432)&
205, 204 : 94.68 \%\\
\hline 
\end{tabular}\par}
\vspace{0.3cm}

\noindent Table III : The network attempts to learn the monks' problem and illustrate
a case were the naive Bayesian model for classification fails. However, with
noise and with partial independence of attributes, our network obtained good
classification efficiency.
\medskip{}

It may be noted that the success rate in the second dataset is poor compared
to the other two. Also, note that the definition of this dataset is antithetical
to the assumptions of naive Bayesian networks on the clustering of classes in
feature space. This is a clear example where naive Bayesian ideas does not work.
While the success in the first dataset illustrate the power of the proposed
algorithm, it is not clear if noise has any effect on the third dataset. Further
studies are required to explore such possibilities.

\section{Conclusion}

Bayes' rule on how the degree of belief should change on the basis of evidences
is one of the most popular formalism for brain modeling. In most implementations,
the degree of belief is computed in terms of the degree of agreement to some
known criteria. However, this has the disadvantage that some of the minor differences
might be left unnoticed by the classifier. We thus device a classifier that
pays more attention to differences rather than similarities in identifying the
classes from a dataset. In the training epoch, the network identifies the apparent
differences and magnify them to separate out classes. We applied the classifier
on many practical problems and found that this makes sense. To illustrate some
of the features of the network, we discuss four examples from the UCI repository.
The highlights of the features of the network are:

1. In all the examples the classification accuracy in both training and testing
sets are comparable. This means that the network has successfully picked up
the right classification information avoiding any possible overfitting of data.

2. Unlike back propagation or its variant, the network converges to the same
accuracy irrespective of initial conditions. 

3. The training time is less compared to other networks and the accuracy is
better.

4. The network topology may be automatically optimized using parallel computation
and the network is well suited for parallel architecture offering high throughput.

We also point out two areas for future research. One is the implementation of
the network on a parallel virtual machine (PVM) and the second is the effect
of noise on the classifier accuracy.

\end{document}